\documentclass[sigconf, nonacm]{acmart}

\renewcommand\footnotetextcopyrightpermission[1]{}
\setcopyright{none}

\usepackage{makecell}

\def\shorten{\looseness=-1}

\usepackage{listings}
\usepackage{xcolor}  
\usepackage{enumitem}
\usepackage{algorithm}
\usepackage{algpseudocode}
\usepackage[table]{xcolor}
\usepackage{pifont}
\usepackage[normalem]{ulem}

\usepackage{tcolorbox}
\usepackage{fontawesome5}
\definecolor{darkblue}{RGB}{0,0,150}

\let\oldemph\emph
\renewcommand{\emph}[1]{\oldemph{#1}}

\usepackage{enumitem}
\setlist[itemize]{leftmargin=*, nosep}

\newtheoremstyle{bolddef}
  {6pt}
  {6pt}
  {\itshape}
  {0pt}        
  {\bfseries}
  {.}
  {0.5em}
  {}

\theoremstyle{bolddef}
\newtheorem{definition}{Definition}

\usepackage{hyperref}

\hypersetup{
  colorlinks=true,
  linkcolor=black,
  citecolor=black,
  urlcolor=blue
}

\AtBeginEnvironment{thebibliography}{%
  
}

\begin{document}

\title[Is Agent Memory a Database? Rethinking Data Foundations for Long-Term AI Agent Memory]{Is Agent Memory a Database? \\
Rethinking Data Foundations for Long-Term AI Agent Memory}

\author{Abdelghny Orogat}
\affiliation{%
  \institution{Concordia University}
}

\author{Essam Mansour}
\affiliation{%
  \institution{Concordia University}
}


\begin{abstract}
Long-running AI agents need persistent memory.
Memory supports learning across sessions, reduces repeated context injection, and enables auditing of past decisions.
Current agent memory systems and database paradigms treat memory as storage. They localize correctness at records, embeddings, or edges. Each supplies only some of the capabilities that long-term memory requires. The result is four recurring failure modes: unregulated growth, missing semantic revision, capacity-driven forgetting, and read-only retrieval. In our vision, long-term agent memory is a new data-management workload. Its correctness is a property of the state trajectory, not of individual records.
We formalize this as Governed Evolving Memory (GEM).
GEM replaces record-level database operations with four state-level operators: ingestion, revision, forgetting, and retrieval.
Six correctness conditions govern how the state evolves.
Three structural observations establish that no record-level system can satisfy these conditions, regardless of the storage model. We realize the abstraction in \textsc{MemState}, a prototype on a property-graph backend. \textsc{MemState} validates feasibility and exposes the gap to a native engine. We outline three research directions that define memory-centric data management as a workload.


\end{abstract}

\maketitle

\begin{tcolorbox}[
colback=gray!6,
colframe=black!25,
boxrule=0.5pt,
arc=2pt,
left=3pt,right=3pt,top=2pt,bottom=2pt]
\small
\noindent\textbf{Key Contributions}
\begin{itemize}[leftmargin=1.1em, nosep, topsep=1pt]
\item A data-centric formalization of long-term agent memory as $M_t = (D_t, S_t, P_t)$ with four operators (ingestion, revision, forgetting, retrieval) and six correctness conditions
\item A reframing of agent memory as a new data-management workload, identifying where existing data systems fall short
\item \textbf{MemState}: a prototype built on an embedded property graph engine demonstrating the proposed foundations
\item A research agenda for governed, evolving memory in long-lived AI agents
\end{itemize}
\vspace{0.5mm}
\noindent\textbf{Code:} \faGithub\ \textcolor{darkblue}{\href{https://github.com/CoDS-GCS/MemState}{https://github.com/CoDS-GCS/MemState}}
\end{tcolorbox}

\section{Introduction}
\label{sec:introduction}

\begin{figure}[t]
    \centering
    \includegraphics[width=\linewidth, trim=.0cm 7.cm 11.7cm 3.cm, clip]{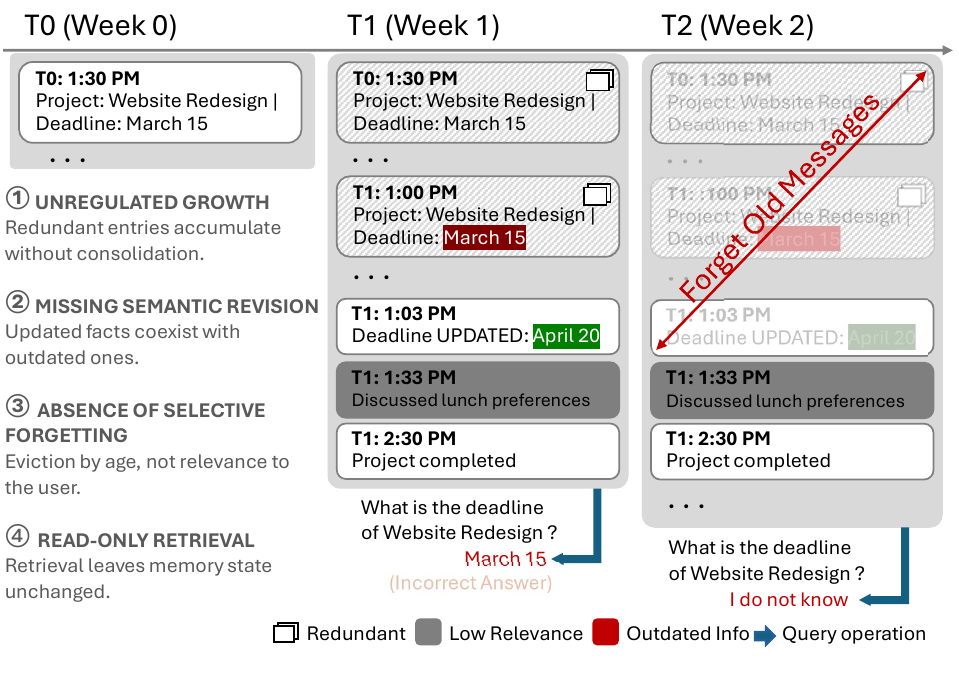}
    \vspace{-9mm}
\caption{Agent memory as an append-only record store over three
weekly snapshots. Built on database operations, it appends new
records and evicts old ones by age, never consolidating the
state. This exposes four failures (\ding{172}, \ding{173}, \ding{174}, and \ding{175}).}
    \vspace{-5mm}
    \label{fig:mem_compare}
\end{figure}

\begin{table*}[t]
\vspace{-2mm}
\centering
\footnotesize
\setlength{\tabcolsep}{4pt}
\renewcommand{\arraystretch}{0.95}
\caption{Coverage of the four capabilities required by governed
memory across database paradigms and agent memory systems.
Each cell describes how the family addresses the capability.
``None'' denotes no native support.
No family supports all four, and each contributes one substrate
strength the others lack.}
\label{tab:limitations}
\vspace{-4mm}
\begin{tabular}{l|p{2.cm}|p{2.4cm}|p{1.6cm}|p{1.9cm}|l}
\toprule
\textbf{Family} &
\makecell{\textbf{Relevance-driven}\\\textbf{retention}} &
\makecell{\textbf{Dependency-aware}\\\textbf{propagation}} &
\makecell{\textbf{Graded}\\\textbf{attenuation}} &
\makecell{\textbf{State-modifying}\\\textbf{retrieval}} &
\textbf{Substrate strength} \\
\midrule
\multicolumn{6}{l}{\textit{Database paradigms}} \\
\midrule
Relational &
None &
Foreign key only &
None &
None &
Schema, ACID \\
Key-value / Document &
None &
None &
None &
None &
Flexible or no schema \\
RDF / Property Graph &
None &
Typed, entity-grain &
None &
None &
Typed structural relations \\
Temporal DB &
None &
None &
None &
None &
Versioned histories \\
Vector DB &
None &
Geometric proximity &
None &
None &
Semantic similarity \\
\midrule
\multicolumn{6}{l}{\textit{Agent memory systems}} \\
\midrule
Tiered (MemGPT, MemOS) &
None &
None &
None &
None &
Two-level paging \\
Fact-extraction (Mem0) &
None &
None &
None &
None &
Atomic fact maintenance \\
Graph-structured (Zep) &
None &
Single-edge invalidation &
None &
None &
Bi-temporal edges \\
Consolidation-based (MIRIX, EverMemOS) &
None &
None &
\textit{Foresight expiry} &
None &
Typed components, scene consolidation \\
RL-driven (Mem-$\alpha$, Memory-R1) &
None &
None &
None &
None &
Learned update policies \\
\midrule
Generative Agents &
\textit{Importance ranking} &
None &
None &
\textit{Recency on access} &
Importance + reflection \\
\bottomrule
\end{tabular}
\vspace{-2mm}
\end{table*}

AI agents operate as persistent systems that interact with
users, tools, and environments~\cite{crewai2025,
openaiagents2025, langgraph2025}.
Unlike question answering systems~\cite{omar2023universal,
omar2025chatty}, they must maintain and revise information
across sessions~\cite{tan2025membench, hu2025evaluating}.
To support this behavior, agents persist information in
external memory beyond the context
window~\cite{packer2023memgpt, li2025memos, xu2025amem,
wang2025mirix, rasmussen2025zep}.
This persistent state shapes whether agent behavior remains stable as interactions accumulate or its performance degrades~\cite{wulongmemeval, hu2025evaluating}.
Long-term memory therefore changes what agents can do.
It lets agents learn across tasks and sessions by carrying
prior decisions and constraints forward.
It also reduces inference cost and latency by avoiding
repeated context injection.

Current memory designs do not preserve these properties.
Most instead follow an accumulation strategy that continuously
appends new information while leaving stored entries
unchanged~\cite{langgraph2025, openaiagents2025,
wang2025mirix, xu2025amem, rasmussen2025zep}.
Figure~\ref{fig:mem_compare} traces this behavior over three
weekly snapshots of the same memory.
Each column represents the memory state at one week, and each
box represents a stored entry.
From Week~0 to Week~1, new records are added and redundant
entries accumulate.
As the state grows, updated facts remain beside obsolete ones.
By Week~2, older entries are evicted by age rather than by
their importance to the user.
Retrieval then operates over this evolving clutter.
These behaviors already appear in daily LLM applications.
ChatGPT~\cite{Chatgpt} and Claude~\cite{claude} retain user
preferences yet still surface outdated facts as current.
Cursor~\cite{cursor} and Claude Code~\cite{claudecode} learn a
codebase yet lose earlier decisions as context grows.
The cost falls on the user.
Users re-explain context the system has already seen.
They pay rising inference cost as context grows.

These systems inherit record-level CRUD operations
(create, read, update, delete) from traditional databases.
As a result, memory operations act on individual records
rather than on the evolving memory state itself.
This mismatch produces the four recurring failure modes shown
in Figure~\ref{fig:mem_compare}.

\noindent\textbf{\ding{172}~\emph{Unregulated growth.}}
Append-only ingestion accumulates redundant and low-relevance
entries.
When the user re-explains a task, the same facts are ingested
again (e.g., the project record \uline{``Website Redesign $|$
Deadline: March~15''} is stored twice at Week~1).
These redundant entries compete at retrieval time and consume
LLM context window space, crowding out useful content.

\smallskip
\noindent\textbf{\ding{173}~\emph{Missing semantic revision.}}
Updates are appended rather than integrated into existing
entries.
At Week~1, \uline{``Deadline UPDATED: April~20''} is stored as a new
entry while ``Deadline: March~15'' remains.
A query ``What is the deadline of Website Redesign?'' may
return ``March~15'' instead of ``April~20''
(based on semantic similarity between the query and message
embeddings)~\cite{tan2025membench, hu2025evaluating}.



\smallskip
\noindent\textbf{\ding{174}~\emph{Absence of selective forgetting.}}
Memory must evict content as storage fills.
But eviction is driven by age or capacity rather than by
importance to the user.
At Week~2 in Figure~\ref{fig:mem_compare}, the project deadline
is evicted while the low-relevance entry
\uline{``Discussed lunch preferences''} persists.
The same deadline query now returns ``I do not know,'' even
though the user asked it before.
The system cannot retain facts according to their relevance to
the user~\cite{maharana2024locomo}.

\smallskip
\noindent\textbf{\ding{175}~\emph{Read-only retrieval.}}
Retrieval returns facts but never updates the memory
state~\cite{wulongmemeval}.
The user queries the project deadline every week, yet that
entry gains no importance and is later evicted like any other (Failure \ding{174}).
User interaction patterns cannot reinforce useful information
or protect it from forgetting.
Frequently accessed facts therefore compete with stale content
on equal terms.

\smallskip
These limitations reflect an abstraction gap, not
implementation issues.
Each failure traces to one CRUD operation: create cannot
integrate, update cannot propagate, delete cannot regulate
relevance, and read cannot adapt.
Larger context windows or better retrieval do not resolve this
mismatch~\cite{tan2025membench, hu2025evaluating}.
The limitation lies in missing evolution semantics, not in
retrieval quality.

\smallskip
\noindent\textbf{Contributions.}
This paper positions long-term agent memory as a new 
data-management workload whose correctness lives in the state 
trajectory, not in individual records.
Our contributions are:

\begin{itemize}
\item A \textbf{four-capability analytical lens}
(relevance-driven retention, dependency-aware propagation,
graded attenuation, and state-modifying retrieval) showing
that no database paradigm or agent memory system supplies all
four (Section~\ref{sec:why_fail}).

\item \textbf{Governed Evolving Memory (GEM)}, a state
abstraction that replaces record-level CRUD with four
state-level operators (ingestion, revision, forgetting,
retrieval) and defines six correctness conditions over the
state trajectory.
Three structural observations show that no CRUD-based system
can satisfy these conditions, regardless of substrate
(Section~\ref{sec:formalization}).

\item \textbf{\textsc{MemState}},
a prototype that realizes GEM on a property-graph backend
with topic-based storage, typed dependencies, and declarative
policies.
The prototype validates feasibility and exposes what a native
engine must provide (Section~\ref{sec:proposed}).

\item A \textbf{research agenda} 
of three directions
covering a native engine, trajectory-level correctness, and privacy under multi-tenant memory, with explicit success criteria. 
(Section~\ref{sec:agenda}).
\end{itemize}
\section{Why Current Abstractions Fail}
\label{sec:why_fail}
We examine database paradigms and recent agent memory systems through
four capabilities required by governed memory.
Each is a column in Table~\ref{tab:limitations} and a failure
mode in Figure~\ref{fig:mem_compare}.
Table~\ref{tab:limitations} covers five database paradigms and
five families of agent memory systems, plus Generative Agents~\cite{park2023generative} as
the closest single approach.
Each contributes one substrate strength.
None covers all four capabilities.

\smallskip
\noindent\textbf{Database paradigms and memory families.}
Database paradigms differ by what they store and how they update.
Relational stores manage records under a fixed
schema~\cite{codd1970relational}.
Key-value and document stores relax this, using no schema or a
flexible one~\cite{decandia2007dynamo,chang2008bigtable}.
RDF and property graphs add typed relations between
entities~\cite{angles2008survey,francis2018cypher,
neumann2010x,perez2009semantics}.
Temporal databases version tuples to preserve
history~\cite{jensen2002temporal,snodgrass1999developing}.
Vector databases index embeddings for semantic
similarity~\cite{wang2021milvus,pan2024survey}.

Agent memory systems group into five families by their primary
mechanism.
Tiered designs (MemGPT~\cite{packer2023memgpt}, MemOS~\cite{li2025memos}) implement
two-level paging: a small in-context active tier and a larger external storage tier.
Evicting from the active tier by age or size when it fills.
Fact-extraction systems (Mem0~\cite{chhikara2025mem0}) parse interactions into atomic
facts and overwrite on conflict.
Graph-structured systems (Zep~\cite{rasmussen2025zep}) link entries via typed edges and
invalidate them bi-temporally to preserve history.
Consolidation-based systems (MIRIX~\cite{wang2025mirix}, EverMemOS~\cite{evermemos})
route content into specialized memory types and cluster related entries into
higher-level structures.
RL-driven systems (Mem-$\alpha$~\cite{wang2025mem}, Memory-R1~\cite{memory_r1}) learn
update policies by reinforcement, rewarding operations by downstream answer quality.
Generative Agents~\cite{park2023generative} rank memories by importance and update
recency on read.
Each family contributes one substrate strength; Table~\ref{tab:limitations} shows where
each falls short.
The rest of this section examines each capability in turn.

\vspace{-2mm}
\subsection{Relevance-Driven Retention} \label{sec:fail_growth}

Relevance-driven retention bounds active memory by utility, stabilizing inference cost as interactions grow. Without it, memory grows monotonically and redundant entries crowd out useful ones (failure mode~\ding{172}).

\smallskip \noindent\textbf{Database paradigms} cap growth by capacity or time, not utility. Relational and document stores use TTL expiry or manual delete~\cite{elmasri2016fundamentals}; temporal databases enforce retention windows on versioned tuples~\cite{jensen2002temporal}; vector databases prune embeddings by capacity or age~\cite{wang2021milvus,pan2024vector}. High- and low-utility facts age out equally (in Figure~\ref{fig:mem_compare}, \uline{``Discussed lunch preferences''} expires on the same schedule as the project deadline). No paradigm bounds memory by relevance.

\smallskip
\noindent\textbf{Agent memory systems} repeat the pattern.
Tier-based, lifecycle-based, capacity-based, and
learned-controller eviction all key on age or
size~\cite{packer2023memgpt,li2025memos,
evermemos,chhikara2025mem0,wang2025mem,memory_r1}.
Generative Agents~\cite{park2023generative} come closest.
They combine an importance score with recency decay to rank
observations at retrieval.
Both are local heuristics applied at retrieval time, not policies
that bound the active footprint.
Memory grows because low-importance observations are never
attenuated.

\vspace{-2mm}
\subsection{Dependency-Aware Propagation} \label{sec:fail_revision}

Dependency-aware propagation keeps related facts consistent when one changes. The agent then reasons over coherent state rather than divergent values. Without it, contradictions yield wrong answers (failure mode~\ding{173}).

\smallskip \noindent\textbf{Database paradigms} offer two relevant techniques, neither propagating along semantic dependencies. Active databases~\cite{widom1996active,ceri1991deriving} propagate via ECA rules over flat relational state, so updates are referential, not content-aware. Materialized view maintenance~\cite{olteanu2024recent} propagates along fixed view schemas, not evolving semantic units. Elsewhere, append-only relational and vector stores return outdated and current values~\cite{codd1970relational,wang2021milvus,pan2024survey}; in-place updates destroy the evidence chain~\cite{decandia2007dynamo,chang2008bigtable}; temporal stores return superseded values as current~\cite{jensen2002temporal,snodgrass1999developing}; and property graph or RDF stores desynchronize at entity grain~\cite{angles2008survey,francis2018cypher,feng2023kuzu,neumann2010x,perez2009semantics}. No DBMS re-evaluates dependent facts when one fact changes.

\smallskip \noindent\textbf{Agent memory systems} inherit the gap. Mem0~\cite{chhikara2025mem0} overwrites or deletes on conflict, destroying the evidence chain. Zep~\cite{rasmussen2025zep} invalidates edges bi-temporally. It is the strongest mechanism but operates one edge at a time. In Figure~\ref{fig:mem_compare}, invalidating the deadline edge at Week1 does not re-evaluate edges to team assignments or meetings. Dependencies then drift silently. Consolidation-based systems~\cite{wang2025mirix,evermemos} overwrite or re-cluster at component grain, not along dependencies. RL-driven systems~\cite{wang2025mem,memory_r1} update without dependency structure.

\vspace{-1mm}
\subsection{Graded Attenuation} \label{sec:fail_forget}

Graded attenuation deprioritizes obsolete content while preserving history for audit, unlike tiering, which moves whole entries by capacity. Without it, obsolete entries compete with current ones (failure mode~\ding{174}).

\smallskip \noindent\textbf{Database paradigms} attenuate by time or capacity. TTL, retention windows, and version pruning act on age and size, not importance~\cite{jensen2002temporal,snodgrass1999developing,wang2021milvus}. Removal is binary. No paradigm demotes content while keeping it recoverable. In Figure~\ref{fig:mem_compare}, the high-utility project deadline ages out while a low-relevance entry survives.

\smallskip \noindent\textbf{Agent memory systems} evict by FIFO, lifecycle stage, or store size~\cite{packer2023memgpt,li2025memos,evermemos}, or delete locally~\cite{chhikara2025mem0}, without relevance. EverMemOS~\cite{evermemos} is the partial exception. It expires time-bounded foresight at retrieval while keeping the rest. Zep~\cite{rasmussen2025zep} caches ingestion-time summaries that aid retrieval but can inflate the store~\cite{chhikara2025mem0}. Generative Agents~\cite{park2023generative} rank observations at retrieval by recency, importance, and relevance but never attenuate them from the store.

\subsection{State-Modifying Retrieval} \label{sec:fail_retrieval}

State-modifying retrieval updates salience on each read so important content stays prominent and stale content fades. Without it, accessed and stale facts compete equally (failure mode~\ding{175}).

\smallskip \noindent\textbf{Database paradigms} treat retrieval as a pure read. A query returns content and leaves state unchanged. Recent retrieval indexes scale read-only access~\cite{malkov2018efficient} but do not let a read update the state it reads.

\smallskip \noindent\textbf{Agent memory systems} do the same. In Figure~\ref{fig:mem_compare}, the Week1 deadline query returns the stored value without reinforcing it, propagating access, or shaping future retrievals. Zep\cite{rasmussen2025zep} reranks by mention frequency, a read-time heuristic, not a committed state change. Generative Agents~\cite{park2023generative} update a memory's recency on access, but only as an embedded heuristic with no correctness condition. Governed memory needs retrieval as a first-class operator that returns output and a state transition (Section~\ref{sec:formalization}).

\smallskip

\smallskip

\noindent\textbf{$\star$ Is Agent Memory a Database?}
\noindent No. CRUD governs records; correctness here lives in
the state trajectory. Agent memory is a new data-management
workload, not a database problem.

\section{Governed Evolving Memory}
\label{sec:formalization}

\begin{figure}[t]
\vspace{-2mm}
    \centering
    \includegraphics[width=\linewidth, trim=.2cm 8.cm 9.5cm 3.cm,
        clip]{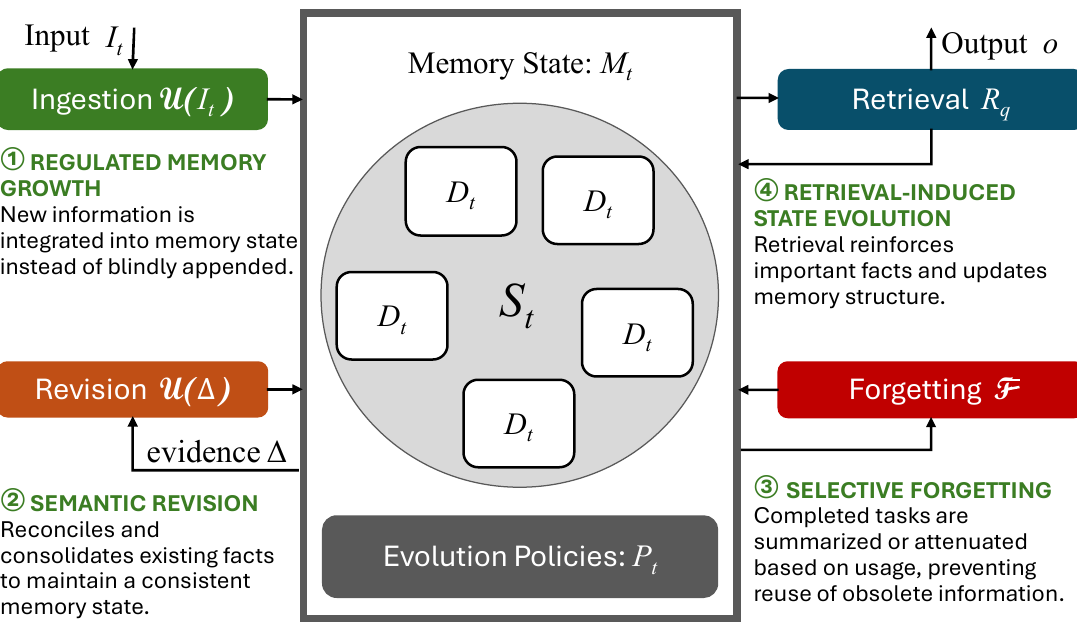}
    \vspace{-6mm}
    \caption{Our GEM Abstraction. The state
$M_t = (D_t, S_t, P_t)$ holds semantic units~($D_t$), their
structural organization~($S_t$), and declarative evolution
policies~($P_t$). These three elements must be explicit in any
compliant implementation; their realization varies by backend.
Four state-level operators replace record-level CRUD:
ingestion, revision, forgetting, retrieval.}
    \label{fig:mem_state}
    \vspace{-2mm}
\end{figure}

This section introduces our Governed Evolving Memory (GEM), the
abstraction that supplies the four capabilities.
Memory is a global state that evolves through structured
operations rather than append-only accumulation.
Figure~\ref{fig:mem_state} illustrates GEM.

\vspace{-1mm}
\subsection{Memory State}
\label{sec:mem_rep}


Three elements must be explicit in the data model.
Content $D_t$ specifies what is stored.
Structure $S_t$ specifies how stored elements connect.
Policies $P_t$ specify how the state is allowed to change.
If any element remains implicit, evolution semantics cannot be
enforced.
A \emph{semantic unit} is the atom of $D_t$, carrying a value
history and a salience signal.
Its boundary is a design decision.
Finer units scatter related facts across
boundaries, requiring graph traversal at retrieval and
risking partial recall.

\begin{definition}[\textbf{Governed Evolving State}]
At time $t$, an agent's memory state is the tuple
$M_t = (D_t, S_t, P_t)$, where $D_t$ denotes stored content
organized as semantic units that hold all related data elements, $S_t$ denotes the structural
organization over that content, and $P_t$ denotes policies
governing access, ingestion, revision, and forgetting.
\end{definition}

\begin{definition}[\textbf{Memory Evolution}]
Memory evolves according to a state transition function
$M_{t+1} = \mathcal{U}(M_t, I_t, R_t)$, where $I_t$ is new
external input and $R_t$ is an internal operation that may
alter state.
The sequence $\{M_t\}_{t \ge 0}$ is the memory trajectory.
\end{definition}

\begin{definition}[\textbf{Evolution Policies}]
\label{def:policies}
$P_t$ is a set of typed rules
$\langle$\emph{event}, \emph{condition}, \emph{action}$\rangle$,
where \emph{event} identifies the operation that triggers
evaluation, \emph{condition} is a predicate over $M_t$, and
\emph{action} is a state-level transition.
Policies are declarative: they specify what transitions occur
and when, independent of how operators execute them.
\end{definition}

\subsection{State-Level Operators}
\label{sec:mem_op}

Four operators act over global memory under policy constraints,
each supplying one capability from Section~\ref{sec:why_fail}.

\smallskip
\noindent\textbf{Ingestion} integrates input $I_t$ into the
existing state by producing
$M_{t+1} = \mathcal{U}(M_t, I_t, \emptyset)$ under constraints
in $P_t$.
An updated value is recorded against the existing semantic
unit; the prior value is retained as historical evidence.

\smallskip
\noindent\textbf{Revision} produces
$M_{t+1} = \mathcal{U}(M_t, \emptyset, \textsc{rev}(\Delta))$
from internal evidence $\Delta$ under $S_t$ and $P_t$.
It reconciles overlapping units, propagates updates along
$S_t$, and preserves superseded values with provenance.

\smallskip
\noindent\textbf{Forgetting} is a policy-governed transition
$M_{t+1} = \mathcal{U}(M_t, \emptyset, \mathcal{F})$, where
$\mathcal{F}$ regulates the influence of stored content by
relevance signals without destructive deletion.
Content in $D_t$ carries salience signals that rise on access
and decay on disuse, at sub-unit granularity, so part of a
unit may be attenuated while the rest stays current.
Attenuation runs as a ladder from partial compression to full
archiving, so the operator is graded rather than binary.

\smallskip
\noindent\textbf{Retrieval} maps query $q$ to output $o$ via
$\mathcal{R}(M_t, q) \to o$ and induces
$M_{t+1} = \mathcal{U}(M_t, \emptyset, \mathcal{R}_q)$
under $P_t$.
Every read updates the salience of accessed units, so retrieval
is a state transition rather than a read-only operation.

\subsection{Memory Correctness}
\label{sec:correctness}

Correctness cannot be defined at the level of individual
records, because contradictory or outdated entries may coexist
within $D_t$.
Correctness is a property of the trajectory along three axes:
what queries return, what the state preserves, and how the
state adapts.

\begin{definition}[\textbf{Memory Correctness}]
\label{def:correctness}
Let $u_i \in D_t$ be a semantic unit with a value history
$V_i = \langle (v_1, t_1, \pi_1), \ldots, (v_k, t_k, \pi_k)
\rangle$, where each entry records a value $v$, a timestamp $t$,
and provenance $\pi$.
A memory system is \emph{correct} if the following six
conditions hold.

\smallskip
\noindent\textbf{C1 (Query soundness).} The response
$\mathcal{R}(M_t, q) \to o$ reflects the most recent
non-archived value $v_k$ as current; prior values appear only
when $q$ explicitly requests historical context.

\smallskip
\noindent\textbf{C2 (Transition soundness).} Every transition
$M_t \to M_{t+1}$ respects $P_t$, and no revision produces a
state in which a superseded value is returned as current.

\smallskip
\noindent\textbf{C3 (Dependency consistency).} For every pair
$(u_i, u_j)$ connected by a typed edge $e \in S_t$ with
propagation semantics, an update to $u_i$ triggers evaluation
of $u_j$ under $P_t$.

\smallskip
\noindent\textbf{C4 (Provenance preservation).} Forgetting and
revision preserve the provenance chain of any unit that remains
reachable.

\smallskip
\noindent\textbf{C5 (Bounded active state).} For every
interaction count $n$, the active memory satisfies
$|D_t^{\text{active}}| \le \beta(n)$ for a policy-defined bound
$\beta$; archived content remains recoverable.

\smallskip
\noindent\textbf{C6 (Retrieval-induced adaptation).} Every
retrieval that accesses $u_i$ induces a transition in which the
salience of $u_i$ is updated.
Repeated retrieval strictly reduces $u_i$'s eligibility for
attenuation.
\end{definition}

C1--C2 govern what queries return.
C3--C4 govern what the state preserves.
C5--C6 govern how the state adapts.
They define correctness as a property of
$\{M_t\}_{t \ge 0}$, not of any individual record.
Every failure mode in Figure~\ref{fig:mem_compare} reduces to
the violation of at least one.\shorten

\subsection{Three Structural Observations}
\label{sec:crud_insufficient}

The four operators are not a recombination of CRUD:

\smallskip
\noindent\textbf{Observation 1 (Retrieval).}
A pure-function retrieval operator cannot satisfy C6.
C6 requires retrieval itself to induce a state transition.
Caches, materialized views, and post-retrieval triggers can
record that an access occurred, but they cannot lift the
operator out of being a pure function, because the
state-modifying step is decoupled from the query.

\smallskip
\noindent\textbf{Observation 2 (Forgetting).}
C5 in its relevance-driven form is jointly unenforceable with
C6 above any CRUD-based engine.
If retrieval is read-only, relevance must be approximated by
external signals, and capacity-driven or time-based attenuation
can bound size but not relevance.

\smallskip
\noindent\textbf{Observation 3a (Ingestion).}
Append-only storage without semantic units cannot satisfy C2.
Two appended values for the same fact coexist with equal
status, and a default query has no engine-level mechanism to
select between them.

\smallskip
\noindent\textbf{Observation 3b (Revision).}
Untyped propagation cannot satisfy C3.
Updates propagate along exact-match references or untyped
edges, so dependencies that the abstraction requires to
re-evaluate are not visible to the engine.

\smallskip
\noindent
These are structural claims, not theorems.
Their consequence: governed memory requires four state-level
operators inside the data model, with $P_t$ checked at commit
and retrieval treated as a write.

\section{Realizing GEM in MemState}
\label{sec:proposed}

We instantiate GEM (Section~\ref{sec:formalization}) in our
prototype, \textsc{MemState}\footnote{Code:
\url{https://github.com/CoDS-GCS/MemState}.}.
The data model supplies C1, C3, and C4 by construction and
lifts C2 to a data-model guarantee.
The four operators of Section~\ref{sec:mem_op} run over this
data model and supply C5 and C6 under transitions.



\vspace{-2mm}
\subsection{Data Model}
\label{sec:mem_rep_impl}
\begin{figure}[t]
\vspace{-2mm}
    \centering
    \includegraphics[width=\linewidth, trim=.cm 8.7cm 8.5cm 3.cm, clip]{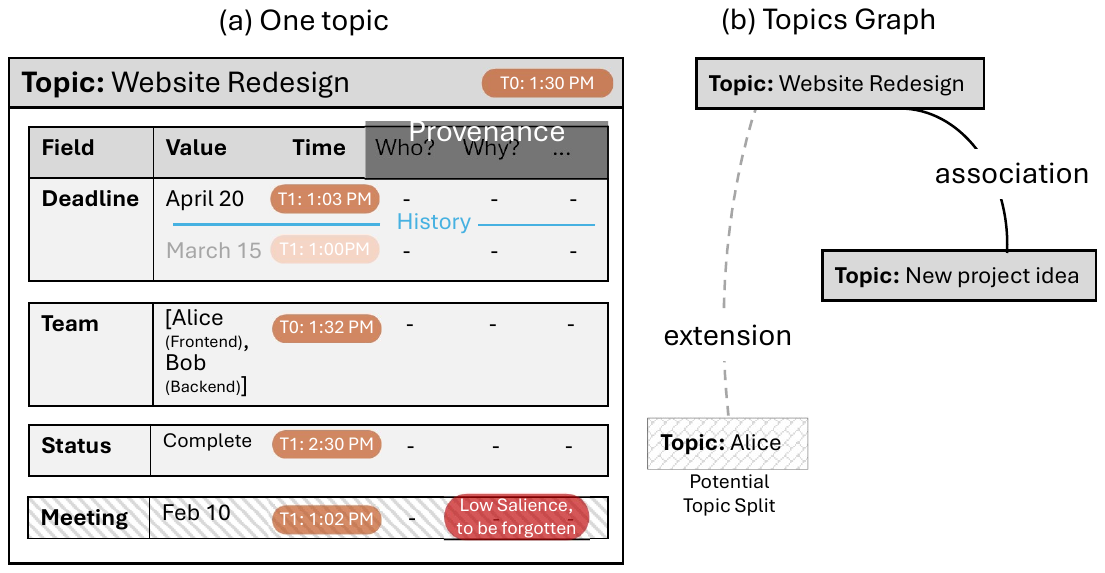}
    \vspace{-7mm}
   \caption{MemState data model. (a) A self-contained topic stores
fields, values, histories, and provenance; low-salience content
can be forgotten. (b) Topics form a typed graph through
\emph{association} and \emph{extension} edges; revision may
promote subsets (e.g., \emph{Alice}) into new topics.}
    \vspace{-1mm}
    \label{fig:proposal}
\end{figure}

MemState realizes $M_t = (D_t, S_t, P_t)$ as an evolving graph
of \emph{topics} on Kuzu~\cite{feng2023kuzu}, an embedded
property graph engine.
Each topic is a self-contained semantic unit with a title, a
summary, a dense embedding, and a set of fields with value histories
(Figure~\ref{fig:proposal}(a)).

\smallskip
\noindent\textbf{Self-contained topics with field histories
(C1, C4).}
A topic groups all fields of one concept in a single unit;
entity-grain designs (Zep) scatter attributes across
nodes, requiring multiple accesses to reconstruct one concept.
Each field is maintained as a history $H_{i,j} = \langle (v, t, \pi) \rangle$:
updates append a new entry rather than overwrite.
Default retrieval returns only $v_k$ (C1); explicit temporal
queries can read any $v_j$ and its provenance (C4).
Once sufficient knowledge accumulates around a subset of
fields, revision promotes it into a standalone topic:
\emph{Alice} splits out of Website Redesign once enough
interactions reference her directly
(Figure~\ref{fig:proposal}(a,b)).

\noindent\textbf{Typed edges in $S_t$ (C3).}
$S_t$ distinguishes two edge types (Figure~\ref{fig:proposal}(b)).
\emph{Extension} edges connect topics where a change in one
can entail a change in the other; a deadline change on
\emph{Website Redesign} may entail a milestone reschedule.
\emph{Association} edges connect related but independent
topics.
C3 propagation must follow entailment, not relatedness,
so revision traverses extension edges only.
Association edges support retrieval context expansion without
propagation.

\noindent\textbf{Declarative policies $P_t$ in state (C2).}
Policies live inside $M_t$ as
$\langle$\emph{event}, \emph{condition}, \emph{action}$\rangle$
rules whose conditions reference $M_t$ directly
(Definition~\ref{def:policies}).
The postcondition on a proposed $M_{t+1}$ is evaluated against
$P_t$ before commit, and a violating transition is rejected.
This lifts C2 to a data-model-level guarantee, closing Observation~3a.
Listing~\ref{lst:policies} shows a representative policy.
Policies can be added, modified, or replaced without changing
operator code.

\vspace{-6mm}
\subsection{Operators on the Kuzu Substrate}
\label{sec:mem_op_impl}

State evolution is realized through the four operators of
Section~\ref{sec:mem_op}.
Ingestion and retrieval are client-facing and synchronous.
Revision and forgetting run asynchronously as policy-triggered
maintenance.
Algorithm~\ref{alg:transition} shows their unified template on
Kuzu.
An incoming event is dispatched to one of four operator
branches (lines 3--12), each writing through field-level,
topic-level, and graph-level primitives on the property graph.
$P_t$ is then evaluated against the proposed $M_{t+1}$ and the
transition either commits atomically or aborts (line 14).
The atomic commit is the mechanism that lifts C2 to a
data-model-level guarantee.
The salience increment inside the retrieval branch (line 11) is
the mechanism for C6.\shorten

\begin{lstlisting}[
  float=t,
  caption={A representative MemState policy: when a field
changes, mark dependent topics for revision (C3).},
  label={lst:policies},
  basicstyle=\small\ttfamily,
  keywordstyle=\bfseries,
  morekeywords={POLICY, ON, WHEN, DO, WITH, EXISTS},
  mathescape=true,
  frame=single,
  framesep=4pt,
  columns=fullflexible,
  keepspaces=true,
  xleftmargin=4pt,
  aboveskip=4pt,
  belowskip=4pt
]
POLICY propagate-on-change
  ON   field_updated
  WHEN EXISTS dependent_topic
  DO   flag_for_revision(dependent_topic)
  WITH evidence = {updated_field, timestamp}
\end{lstlisting}

\setlength{\textfloatsep}{6pt}      
\setlength{\floatsep}{6pt}          
\setlength{\intextsep}{6pt}         

\begin{algorithm}[t]
\setlength{\abovecaptionskip}{0pt}
\setlength{\belowcaptionskip}{0pt}
\caption{GEM transition on a property-graph substrate.}
\label{alg:transition}
\small
\begin{algorithmic}[1]
\Require Memory state $M_t = (D_t, S_t, P_t)$;
         event $e \in \{I_t, q, \Delta\}$
\Ensure Updated state $M_{t+1}$; output $o$ if $e = q$
\State $\mathit{op} \gets \mathit{dispatch}(e)$
       \Comment{ingest, revise, forget, retrieve}
\State \textbf{begin transaction}
\If{$\mathit{op} = \mathit{ingest}$}
  \State LLM picks host topic $\tau$; for each fact $(f,v,t,\pi)$,
         append to $H_{\tau,f}$ or create $f$;
         refresh $\tau$'s embedding; flag extension-linked topics
\ElsIf{$\mathit{op} = \mathit{revise}$}
  \State apply repair (conflict, merge, propagate) for each
         evidence $\delta \in \Delta$
\ElsIf{$\mathit{op} = \mathit{forget}$}
  \State attenuate each $u$ with $\mathrm{salience}(u) < \theta_{*}$
         (compress, hide, archive)
\ElsIf{$\mathit{op} = \mathit{retrieve}$}
  \State route $q$; read selected units; build $o$;
         increment salience of accessed units
         \Comment{C6}
\EndIf
\State evaluate $P_t$ on proposed $M_{t+1}$;
       \textbf{commit} if all postconditions hold, else
       \textbf{abort}
       \Comment{atomic commit lifts C2}
\State \Return $M_{t+1}$ (and $o$ if applicable)
\end{algorithmic}
\end{algorithm}

\noindent\textbf{Ingestion.}
The LLM reads topic titles and summaries to select a host
topic, then reads the topic schema to place the new value
in an existing or new field.
For the deadline update from Figure~\ref{fig:mem_compare}, the
new entry $(\textit{April 20}, T_1, \pi_1)$ is appended to
$H_{\textit{deadline}}$ while
$(\textit{March 15}, T_0, \pi_0)$ remains.
The embedding is refreshed and extension-linked topics are
flagged for revision.
C1, C2, and C4 hold by construction.

\noindent\textbf{Revision.}
Revision detects evidence items in $M_t$ (duplicate topics,
conflicting field values, schema drift, dependency
inconsistencies) and applies the corresponding repair.
Dependency repair walks extension edges, halting at any topic
whose policy condition does not fire. Topic granularity keeps
this frontier far smaller than an entity-grain graph (Zep),
so the walk terminates in few hops. This supplies the
dependency-aware propagation capability of
Section~\ref{sec:fail_revision} at the global level.
Conflict resolution marks superseded values with their
provenance rather than deleting them, preserving C4.

\noindent\textbf{Forgetting.}
Each field maintains a salience score that rises on access and
decays on disuse.
Three thresholds define a graded ladder.
Below $\theta_{\mathit{summary}}$, a history is compressed.
Below $\theta_{\mathit{remove}}$, a field is hidden from active
retrieval.
Below $\theta_{\mathit{archive}}$, the topic is archived but
remains recoverable through explicit lookup.
Retention is driven by salience rather than age or capacity,
satisfying C5.

\noindent\textbf{Retrieval} routes a query into topic-based,
temporal, or structural mode.
In topic-based mode, it inspects titles and summaries, selects
candidate topics, reads schemas, and reads required field
values.
For \emph{``What is the deadline for the Website Redesign?''}
it returns the current \emph{Deadline} value.
The salience increment on every accessed unit is part of the
operator semantics, closing Observation~1.

\vspace{-3mm}
\subsection{From Prototype to Native Engine}
\label{sec:prototype_scope}

\noindent\textbf{What MemState validates.}
MemState is a feasibility sketch on a property-graph substrate.
GEM is realizable on commodity infrastructure: topics, field
histories, embeddings, and policy postcondition checks all
attach to one Kuzu transaction, so C1, C2, and C4 hold by
construction and C5, C6 hold under transitions.

\noindent\textbf{What MemState exposes.}
The substrate is a compatibility layer, not a native expression
of GEM.
Field histories, propagation-bearing edges, and
postcondition-checked commits are reconstructed from generic
graph primitives, so several optimizations are not directly
expressible.
A native engine would store field histories as first-class
bitemporal attributes, attach propagation semantics to
extension edges in the schema, and compile policy
postconditions into the commit protocol.
Retrieval-induced salience updates would compile into a single
read-modify-write primitive, and relevance-driven forgetting
would be scheduled like index maintenance.
These are not implementation gaps; they are research directions
for a data-management workload that no current engine targets.
\vspace{-2mm}
\section{Research Agenda}
\label{sec:agenda}

This section presents three research directions and the success
criteria of the vision.
The directions follow from GEM and MemState, grounded in
capabilities no current engine supports natively.


\vspace{1mm}
\noindent\textbf{A Native Engine for Governed Memory.}
MemState reconstructs topic records, field histories,
propagation-bearing edges, and policy-checked commits from
generic property-graph primitives, at a compatibility-layer
cost.
A native engine must address three problems.
(i)~\emph{Storage layout:} co-locate topics, field histories,
and embeddings on pages to reduce I/O cost per read, following
the node and neighbor co-location used in graph-based vector
engines~\cite{gaussdb-vector}, adapted to units that carry both
a value history and typed dependencies.
(ii)~\emph{Unified indexing:} jointly support semantic
similarity and history predicates, so vector-based, temporal,
and structural queries route through the same physical
organization without duplicating data.
(iii)~\emph{Retrieval as a write:} C6 requires every read to
update salience, but existing query languages separate reads
from writes~\cite{francis2018cypher,perez2009semantics,deutsch2022gql}
and existing indexes optimize read-only access at
scale~\cite{malkov2018efficient}.
A native engine needs an operator that unifies search,
traversal, temporal lookup, and salience updates, with a buffer
strategy that keeps hot topics and dependency context resident.
First targets: (i) an I/O-aware page layout for topics with
field histories and extension edges, (ii) a joint index over
semantic similarity and history predicates, and (iii) the
consistency cost of retrieval-induced salience updates under
concurrent access.

\vspace{1mm}
\noindent\textbf{Correctness and Evaluation.}
Correctness in GEM is a property of the state trajectory
$\{M_t\}_{t \ge 0}$, not of individual records.
Three sub-problems follow.
(i)~\emph{Trajectory benchmark.}
Current benchmarks measure answer-level recall and exercise C1
only
partially~\cite{maharana2024locomo,wulongmemeval,tan2025membench,hu2025evaluating}.
A system that overwrites history or never forgets can still
score well if its recent answers are correct.
A trajectory benchmark needs ground truth at three levels:
the current value of each unit over time~(C2), the dependent
units that change after each update~(C3), and the active
footprint at each interaction count~(C5).
No existing benchmark provides all three.
(ii)~\emph{Policy language for conflict resolution.}
Existing systems either overwrite on
conflict~\cite{chhikara2025mem0} or invalidate edges through
an LLM gate that silently misses
contradictions~\cite{rasmussen2025zep}.
A declarative policy language must express conflict-resolution
rules over field histories with provenance, exploiting typed
edges rather than LLM gating.
(iii)~\emph{Constrained learned controllers.}
Mem-$\alpha$~\cite{wang2025mem} and
Memory-R1~\cite{memory_r1} learn update policies over flat
fact stores without $S_t$ or field histories.
The open problem is a controller that optimizes within
trajectory-level constraints enforced by the commit protocol.
First targets: (i) a 500-turn adversarial workload scoring
Mem0, Zep, and MemState on answer- and trajectory-level
metrics, (ii) a policy language prototype on
LongMemEval~\cite{wulongmemeval} and
LoCoMo~\cite{maharana2024locomo}, and (iii) an RL controller
trained against C2--C5 constraints.

\vspace{1mm}
\noindent\textbf{Privacy and Multi-Tenancy.}
A third direction concerns shared memory.
Production agents often serve multiple tenants over a common
memory instance, which introduces two problems that do not
arise in single-tenant settings.
(i)~\emph{Retrieval-induced information leakage.}
C6 makes retrieval a write operation.
If tenant $A$'s query reinforces topic $\tau$, the salience
update is committed to $M_t$ as a state transition.
A later query from tenant $B$ surfaces $\tau$ through
similarity ranking because its salience score is high.
The salience signal acts as an information leakage path across
tenant isolation boundaries, even when topic content is
access-controlled.
Existing memory systems treat retrieval as a read-only
operation and do not account for this side effect.
(ii)~\emph{Verifiable erasure under evolving state.}
Privacy regulations require provable removal of a tenant's
data upon request.
GEM makes erasure strictly harder than relational delete,
because a forgetting operator must compose with C4
(provenance preservation) and C6 (derived salience).
A tenant's data shapes provenance chains on other topics and
salience aggregates that influenced other tenants' query
results.
Deleting the base records does not erase these derived
signals.
Data exchange research~\cite{fagin2005data} addresses
cross-boundary consistency but not privacy-preserving
forgetting over an evolving state trajectory.
First targets: (i) the information leakage rate between two
tenants on a shared MemState instance and a retrieval operator
with bounded salience side effects, and (ii) a forgetting
operator that erases derived salience and provenance traces
under C4 and C6.

\vspace{1mm}
\noindent\textbf{Success Criteria.}
The vision succeeds when four conditions hold.
(i) At least one DBMS exposes governed-evolution operators as
first-class primitives, and agent frameworks declare retention
and propagation policies declaratively.
(ii) Standardized trajectory-level benchmarks measure C2--C6
violations across long interaction histories.
(iii) Long-horizon deployments show measurable reductions in
temporal-reasoning errors attributable to GEM-conformant
memory.
(iv) Privacy and forgetting guarantees over evolving memory
become as well-understood as ACID guarantees over transactional
storage.
The arc parallels stream processing, which became a recognized
workload once continuous state and event-time semantics moved
from application code into the data model.
\vspace{-2mm}
\section{Conclusion}
\vspace{-1mm}
\label{sec:conclusion}
Long-term agent memory is the workload behind every persistent
AI agent, but no current system treats it as one.
We argued that its correctness is a property of the state
trajectory, not of individual records.
This paper envisions Governed Evolving Memory, an abstraction
that enforces this property through four state-level operators
and six correctness conditions.
Our MemState prototype realizes the abstraction on a
property-graph substrate and exposes what a native engine must
deliver.
Three directions define memory-centric data management as a
workload: a native engine for governed memory, trajectory-level
correctness and evaluation, and privacy under shared salience.
The vision succeeds when long-term memory joins transactions
and streams as a recognized data-management workload.

\bibliographystyle{ACM-Reference-Format}
\bibliography{all_references}

\end{document}